\documentclass[11pt]{article}

\usepackage[preprint]{acl}

\usepackage{times}
\usepackage{latexsym}
\usepackage[T1]{fontenc}
\usepackage[utf8]{inputenc}
\usepackage{microtype}
\usepackage{inconsolata}
\usepackage{graphicx}

\usepackage{booktabs}
\usepackage{multirow}
\usepackage{array}
\usepackage[table]{xcolor}
\usepackage{makecell}
\usepackage{amsmath}
\usepackage{amssymb}
\usepackage{url}

\definecolor{poscell}{RGB}{222,245,229}
\definecolor{negcell}{RGB}{252,225,225}

\newcommand{\pos}[1]{\cellcolor{poscell}{#1}}
\newcommand{\negcellcmd}[1]{\cellcolor{negcell}{#1}}
\newcommand{\feat}[2]{(\textsc{L}#1,\ #2)}
\newcommand{\comet}{\textsc{Comet}}

\definecolor{basegray}{HTML}{F4F4F4}
\definecolor{ampgreen}{HTML}{E8F5E9}
\definecolor{ablred}{HTML}{FDECEA}

\usepackage{xcolor}
\usepackage{tcolorbox}
\tcbuselibrary{skins,breakable}
\newtcolorbox{outputbox}[2][]{
  colback=#2,
  colframe=black!15,
  boxrule=0.3pt,
  arc=2pt,
  left=4pt,
  right=4pt,
  top=3pt,
  bottom=3pt,
  fonttitle=\bfseries\small,
  fontupper=\small,
  enhanced,
  breakable,
  #1
}

\title{Recurrence Is Not Enough: Causally Validating Multilingual SAE Translation Features in Gemma 2 and 3}

\author{
 \textbf{Giang Son Nguyen\textsuperscript{1,2}} \qquad
 \textbf{Nhi Ngoc-Yen Nguyen\textsuperscript{1,3}} \qquad
 \textbf{Wray Buntine\textsuperscript{1,4}} \qquad
 \textbf{Dung D. Le\textsuperscript{1}}
\\
 \textsuperscript{1}VinUniversity \quad
 \textsuperscript{2}Nanyang Technological University \\
 \textsuperscript{3}University of Science and Technology of China \quad
 \textsuperscript{4}Monash University
\\
\texttt{\{son.ng, nhi.nny, wray.b, dung.ld\}@vinuni.edu.vn}
}

\begin{document}
\maketitle
\begin{abstract}
Sparse autoencoder (SAE) features are increasingly used to explain and steer language-model behavior, but it remains unclear whether a feature found in one language context plays the same causal role when processing prompts in another language. 
We study this question using translation-initiation features \cite{wu2026finding}. 
We reproduce the SAE feature discovery method from \citeauthor{wu2026finding} in Gemma 2 and extend it to multilingual settings that vary prompt language, source language, and target language. We then test whether features that recur across settings affect translation behavior by amplifying or ablating their activations during inference. We also examine whether the method can be applied to Gemma 3.

In both models, we observe an identical finding: although we can find more than 20 features that activate frequently across all discovery settings, causal validation shows that nearly all have small or inconsistent effects. In contrast, one feature -- Gemma 2's (L10, 5717) and Gemma 3's (L20, 2456) -- consistently improves COMET scores when amplified and degrades them when ablated across 23 language settings. These results show that feature recurrence can overstate cross-lingual transfer, while identifying a language-agnostic translation-initiation direction in Gemma 2 and Gemma 3.\footnote{Our code is available at: \url{https://github.com/giangson19/recurrence-is-not-enough}}

\end{abstract}

\section{Introduction}
Mechanistic interpretability seeks internal variables that causally support model behavior \citep{elhage2022toy,olah2020zoom}. Sparse autoencoders (SAEs) are widely used to decompose activations into interpretable feature directions \citep{bricken2023monosemanticity,huben2024sparse}. However, SAE explanations can be brittle: automated interpretability scores may fail on basic sanity checks \citep{heap2026automated}, and semantic-looking features may reflect token artifacts \citep{ronge2026coffee}. These concerns are sharper in multilingual settings, where a feature found in one language context may not implement the same computation elsewhere.

We study this issue through the translation-initiation mechanism of \citet{wu2026finding}. They report that a small set of Gemma SAE features activate at translation-specific prompt positions, and that amplifying or ablating them improves or hurts translation quality. However, their evaluation is English-centric: prompts are in English and translation is English-to-X. It is unclear whether the same directions are causal when the prompt, source, or target language changes. This question connects to broader work on whether multilingual LLMs rely on shared language-agnostic or language-specific representations \citep{wendler2024latent_language,brinkmann2025latent_grammar,schut2025think_english}. Moreover, their results were only reported in one model family (Gemma 2's 2B and 9B variants), raising questions of whether the method can be transferred to other models. 

In this reproduction study, we ask: \emph{do SAE translation features causally transfer across multilingual settings?} We extend the discovery and intervention pipeline of \citet{wu2026finding}. 
We use the same models -- Gemma 2 2B  IT \cite{gemmateam2024gemma2improvingopen} and pre-trained SAEs from the Gemma Scope suite \cite{lieberum2024gemmascopeopensparse} -- but vary both translation pair and prompt language. We first identify features shared across four settings (English-to-Chinese with English prompts, Chinese-to-English with Chinese prompts, and Chinese-to-Japanese with either Chinese or English prompts) then intervene on these features in 23 settings. We deliberately remove English's influence by introducing settings where English is not present in the source sentence, or the prompt instruction, or simply removed entirely. 
We evaluate effects of feature intervention using \comet{} scores \cite{rei-etal-2020-comet}.

Our main finding is unexpectedly sparse. Although we found 22 recurring features across all four discovery settings, \textbf{only one---\feat{10}{5717}---shows robust causal transfer}. Amplifying it promotes translation behavior across diverse language pairs and prompt languages, while ablating it usually degrades quality. This suggests that one SAE direction functions as a language-agnostic translation-initiation switch, while the failure of most overlapping candidates cautions against treating feature recurrence as evidence of mechanism transfer. 

Furthermore, we test whether the discovery pipeline can be applied to another model, specifically using Gemma 3 4B IT \cite{gemmateam2025gemma3technicalreport} and pre-trained SAEs from Gemma Scope 2 \cite{mcdougall2025gemmascope2} of similar sizes to Gemma 2's. Despite differences in model size, architecture and pre-training data to Gemma 2, we observe identical results in Gemma 3: we found 22 recurring features across four discovery settings, yet \textbf{only one -- \feat{20}{2456} -- consistently has improving/degrading effects} under intervention across 23 language settings.\footnote{The identical count of 22 recurring features in both models is coincidental; it arises from independent runs of the same pipeline on different models and SAEs.}

Our contributions are:
\begin{itemize}
    \item We test cross-lingual transferability of SAE-based translation features across prompt languages and translation pairs.
    \item We show that multilingual feature recurrence overestimates causal transferability.
    \item We identify one feature in each model, \feat{10}{5717} in Gemma 2 and \feat{20}{2456} in Gemma 3, whose causal effects transfer robustly across settings.
\end{itemize}

The scope of our reproduction study compared to \citet{wu2026finding} is summarized in Table~\ref{tab:scope}.

\begin{table}[t]
\centering
\footnotesize
\setlength{\tabcolsep}{6pt}
\renewcommand{\arraystretch}{1.35}
\begin{tabular}{@{}>{\raggedright\arraybackslash}p{1.45cm}
                >{\raggedright\arraybackslash}p{1.65cm}
                >{\raggedright\arraybackslash}p{3.45cm}@{}}
\toprule
\textbf{Aspect} & \textbf{\citet{wu2026finding}} & \textbf{Ours} \\
\midrule
Models &
Gemma 2 &
Gemma 2, 3 \\

Discovery settings &
\texttt{en2zh\_en} &
\texttt{en2zh\_en}, \texttt{zh2en\_zh}, 
\texttt{zh2ja\_zh}, \texttt{zh2ja\_en} \\

Test settings &
4; English source/prompt &
23; non-English sources and no-English settings \\
\bottomrule
\end{tabular}
\caption{Comparison of experimental scope.}
\label{tab:scope}
\end{table}

\section{Related Work}

\paragraph{Sparse dictionary learning for interpretability and intervention.}
SAEs decompose model activations into sparse feature activations and decoder directions, often yielding more human-interpretable units than raw neurons \citep{bricken2023monosemanticity,huben2024sparse,templeton2024scaling}. Recent work uses SAEs to study multilingual mechanisms, language-specific features, and feature steering \citep{cho2025multilingual_sae,deng2025language_specific_sae,chou2025causal_language_control}. Our work follows this line but emphasizes causal validation under distribution shift: a feature is only considered transferable if the same intervention direction changes behavior across many languages.

\paragraph{Reliability of mechanistic explanations.}
A growing literature questions whether interpretability methods identify stable mechanisms or produce plausible but brittle descriptions. Prior work highlights interpretability illusions \citep{bolukbasi2021interpretability_illusions}, weak automated explanation metrics \citep{heap2026automated}, and token-driven false positives \citep{ronge2026coffee}. Our results reinforce these concerns: 22 features pass a strict discovery criterion, but only one shows robust causal effects.

\paragraph{Multilingual representations and interpretability.}
Mechanistic interpretability remains largely English-dominant, and relatively little work tests whether interpretability findings retain their meaning or causal role across languages \citep{resck-etal-2025-explainability}. This matters because multilingual models may share representations across languages, but can also exhibit language-specific computation or latent-language effects \citep{wendler2024latent_language,brinkmann2025latent_grammar,schut2025think_english}. Some behavioral control directions, such as refusal, appear to transfer across languages \citep{arditi2024refusal_direction,wang2026refusal}. We build on this line by testing whether a translation-initiation feature remains causal when the prompt, source, and target language change.

\section{Method}

\subsection{Notation}

We denote the features as \texttt{(Layer, Feature Index)}. For example, (L10, 5717) refers to the 5717\textsuperscript{th} SAE feature from the 10\textsuperscript{th} layer.

Moreover, we denote each setting as \texttt{src2tgt\_prompt}, where \texttt{src} is the source language, \texttt{tgt} is the target language, and \texttt{prompt} is the language of the translation instruction. 
We use the language codes \texttt{en}, \texttt{zh}, \texttt{ja}, \texttt{ar}, \texttt{ru}, and \texttt{vi} for English, Chinese, Japanese, Arabic, Russian, and Vietnamese, respectively.

\begin{table}[t]
\centering
\small
\begin{tabular}{lcc}
\toprule
Setting & Gemma 2 & Gemma 3 \\
\midrule
\texttt{en2zh\_en} & 42 & 45 \\
\texttt{zh2en\_zh} & 53 & 104 \\
\texttt{zh2ja\_zh} & 83 & 206 \\
\texttt{zh2ja\_en} & 59 & 65 \\
\midrule
Intersection (all 4) & 22 & 22 \\
\bottomrule
\end{tabular}
\caption{Number of consistent features discovered under each setting.}
\label{tab:consistent_features}
\end{table}

\subsection{Feature discovery}


We follow the three-stage protocol of \citet{wu2026finding} for identifying translation-initiation features and use their published codebase as the foundation for our study\footnote{https://github.com/flamewei123/AAAI26-translation-Initiation-Features}. All thresholds match their published values.

\paragraph{Stage 1: feature discovery.}
For each of three prompt strategies---\texttt{prompt}, \texttt{tgt\_lang}, and \texttt{tgt\_idx}---we run forward passes on the 98 examples with target text appended. We extract nonzero SAE activations at the corresponding boundary token position, retain features that activate in more than 60\% of example-template slots, filter by minimum mean activation greater than 10.0, and keep the top 50 features by mean activation. We then take the union across the three strategies, yielding up to 150 candidate features per setting.

The boundary token position depends on the prompt strategy: \texttt{prompt} uses the token immediately before the source text, \texttt{tgt\_lang} uses the last token of the target-language name in the instruction, and \texttt{tgt\_idx} uses the last token of the appended target text.

\paragraph{Stage 2: PCA-based consistency filtering.}
For each candidate feature, we ablate the feature by setting its activation to zero on each discovery example across all prompt templates. We then compute the intervention-induced delta vector, defined as the change in the hidden state at the last-token position, and apply PCA to the collected delta vectors. Following \citet{wu2026finding}, we retain features for which the first principal component explains at least 95\% of the variance. This filters for features whose interventions induce a consistent residual-stream direction rather than noisy or context-dependent effects.

\paragraph{Stage 3: causal intervention.}
For each surviving feature, we evaluate interventions on the 862 held-out examples. We compare greedy baseline generations with two interventions applied at the last prompt token: ablation, which sets the feature activation to zero ($\alpha=0$), and amplification, which doubles the feature activation ($\alpha=2$). We report COMET changes relative to the unmodified baseline.

\begin{table*}[t]
\centering
\small
\setlength{\tabcolsep}{4pt}
\begin{tabular}{lrrrrrrrr}
\toprule
\multirow{2}{*}{Feature} &
\multicolumn{4}{c}{Amplification $\Delta$\comet{}} &
\multicolumn{4}{c}{Ablation $\Delta$\comet{}} \\
\cmidrule(lr){2-5}\cmidrule(lr){6-9}
& en2zh\_en & zh2en\_zh & zh2ja\_zh & zh2ja\_en
& en2zh\_en & zh2en\_zh & zh2ja\_zh & zh2ja\_en \\
\midrule
\feat{0}{8920} & +0.10 & +0.37 & -0.78 & -0.14 & -0.27 & -0.30 & +0.38 & +0.84 \\
\feat{1}{12054} & -0.11 & +0.16 & \negcellcmd{-1.03} & -0.35 & -0.02 & -0.21 & +0.21 & +0.44 \\
\feat{2}{15089} & +0.06 & -0.29 & -0.92 & +0.31 & +0.08 & -0.20 & +0.44 & -0.18 \\
\feat{6}{13460} & -0.72 & -0.46 & -0.17 & -0.21 & -0.16 & +0.15 & -0.26 & +0.75 \\
\feat{7}{3704} & +0.11 & -0.14 & -0.06 & -0.32 & -0.44 & +0.07 & -0.55 & -0.09 \\
\feat{8}{2024} & -0.14 & -0.10 & -0.62 & -0.08 & +0.41 & +0.56 & +0.63 & +0.28 \\
\feat{9}{54} & -0.89 & -0.03 & -0.71 & -0.75 & -0.02 & +0.16 & -0.32 & +0.97 \\
\feat{10}{3986} & \negcellcmd{-1.16} & -0.50 & \negcellcmd{-1.37} & +0.10 & +0.21 & +0.51 & +0.77 & +0.45 \\
\textbf{\feat{10}{5717}} & \pos{\textbf{+4.61}} & \pos{\textbf{+2.33}} & \pos{\textbf{+5.22}} & \pos{\textbf{+3.66}} & \negcellcmd{\textbf{-5.85}} & -0.49 & \negcellcmd{\textbf{-3.96}} & \negcellcmd{\textbf{-4.79}} \\
\feat{11}{3151} & \negcellcmd{-1.96} & -0.18 & \negcellcmd{-1.60} & -0.09 & \pos{+1.01} & +0.19 & \pos{+1.00} & +0.24 \\
\feat{12}{2620} & \negcellcmd{-1.75} & -0.48 & \negcellcmd{-1.68} & +0.10 & +0.67 & +0.45 & \pos{+1.35} & +0.04 \\
\feat{13}{11256} & \negcellcmd{-1.09} & -0.90 & \negcellcmd{-1.27} & +0.07 & +0.05 & +0.10 & +0.90 & +0.49 \\
\feat{14}{7214} & -0.67 & -0.68 & -0.57 & +0.04 & +0.07 & +0.66 & +0.40 & +0.77 \\
\feat{14}{12319} & -0.45 & -0.04 & +0.04 & +0.81 & -0.29 & -0.32 & +0.12 & -0.08 \\
\feat{15}{8610} & +0.07 & -0.70 & -0.71 & +0.97 & -0.81 & +0.96 & +0.12 & +0.08 \\
\feat{16}{3847} & -0.36 & +0.45 & +0.29 & +0.17 & -0.66 & -0.77 & -0.02 & +0.59 \\
\feat{16}{10480} & +0.28 & -0.36 & +0.44 & \pos{+1.03} & -0.69 & -0.05 & -0.55 & -0.16 \\
\feat{17}{7921} & +0.03 & -0.34 & -0.19 & \pos{+1.57} & \negcellcmd{-1.14} & +0.11 & -0.53 & -0.55 \\
\feat{18}{15394} & +0.05 & -0.46 & -0.22 & +0.62 & -0.62 & +0.30 & -0.23 & -0.11 \\
\feat{24}{8119} & \pos{+1.80} & \negcellcmd{-1.13} & \pos{+1.01} & +0.91 & \negcellcmd{-3.07} & +0.84 & \negcellcmd{-1.89} & \negcellcmd{-1.44} \\
\feat{25}{348} & \negcellcmd{-1.96} & \negcellcmd{-19.72} & \negcellcmd{-4.52} & \negcellcmd{-8.70} & \negcellcmd{-8.82} & \pos{+2.96} & \negcellcmd{-8.74} & \negcellcmd{-12.07} \\
\feat{25}{14186} & \pos{+1.67} & \negcellcmd{-5.73} & \negcellcmd{-3.62} & \negcellcmd{-1.28} & \negcellcmd{-9.12} & \pos{+1.61} & \negcellcmd{-5.72} & \negcellcmd{-7.59} \\
\bottomrule
\end{tabular}
\caption{\textbf{Gemma 2}: causal effects of the 22 features recovered in all four discovery settings. Values are changes in \comet{} relative to baseline; green/red cells mark effects with absolute value $\geq 1.0$. Exactly one feature, \textbf{(L10, 5717)}, shows the expected translation-switch signature (consistent positive amplification, consistent negative ablation); the remaining 21 features survived the same recurrence-based selection in the same settings, yet produce effects that are small, inconsistent in sign across settings, or both. 
Corresponding chrF++ results are reported in Appendix~\ref{app:chrf-evaluation}.
}
\label{tab:shared_features_g2}
\end{table*}

\begin{table*}[t]
\centering
\small
\setlength{\tabcolsep}{4pt}
\begin{tabular}{lrrrrrrrr}
\toprule
\multirow{2}{*}{Feature} &
\multicolumn{4}{c}{Amplification $\Delta$\comet{}} &
\multicolumn{4}{c}{Ablation $\Delta$\comet{}} \\
\cmidrule(lr){2-5}\cmidrule(lr){6-9}
& en2zh\_en & zh2en\_zh & zh2ja\_zh & zh2ja\_en
& en2zh\_en & zh2en\_zh & zh2ja\_zh & zh2ja\_en \\
\midrule
\feat{4}{2491} & -0.21 & -0.40 & +0.13 & -0.83 & +0.67 & +0.21 & +0.58 & +0.86 \\
\feat{6}{40} & +0.29 & -0.02 & +0.05 & +0.18 & +0.09 & -0.08 & +0.08 & +0.01 \\
\feat{6}{236} & +0.22 & -0.07 & +0.42 & +0.15 & -0.17 & -0.14 & +0.37 & +0.15 \\
\feat{11}{6005} & +0.54 & +0.22 & +0.73 & +0.18 & -0.16 & -0.26 & +0.05 & -0.11 \\
\feat{12}{273} & \negcellcmd{-3.14} & \negcellcmd{-1.39} & \negcellcmd{-2.05} & \negcellcmd{-6.41} & \negcellcmd{-19.72} & \negcellcmd{-10.49} & \negcellcmd{-17.94} & \negcellcmd{-18.67} \\
\feat{16}{1562} & +0.40 & +0.22 & +0.38 & -0.29 & +0.17 & -0.22 & +0.23 & -0.40 \\
\feat{16}{4659} & -0.05 & -0.06 & +0.08 & -0.42 & +0.16 & +0.07 & +0.77 & +0.08 \\
\feat{17}{1125} & +0.04 & -0.69 & -0.46 & -0.53 & \negcellcmd{-4.05} & \negcellcmd{-1.07} & \negcellcmd{-2.57} & \negcellcmd{-3.18} \\
\feat{18}{11066} & +0.58 & +0.72 & +0.95 & +0.79 & -0.62 & \negcellcmd{-1.30} & -0.46 & \negcellcmd{-1.95} \\
\feat{19}{383} & +0.10 & +0.66 & \pos{+1.31} & +0.39 & \negcellcmd{-1.12} & \negcellcmd{-1.06} & -0.89 & \negcellcmd{-2.16} \\
\feat{20}{189} & +0.76 & +0.67 & \pos{+1.18} & +0.65 & -0.43 & -0.71 & +0.12 & -0.79 \\
\textbf{\feat{20}{2456}} & \pos{\textbf{+1.40}} & \pos{\textbf{+1.45}} & \pos{\textbf{+2.23}} & \textbf{+0.59} & \negcellcmd{\textbf{-3.29}} & \negcellcmd{\textbf{-3.49}} & \negcellcmd{\textbf{-3.78}} & \negcellcmd{\textbf{-4.15}} \\
\feat{21}{862} & +0.77 & +0.38 & +0.82 & +0.31 & \negcellcmd{-1.04} & -0.51 & -0.19 & \negcellcmd{-1.45} \\
\feat{22}{373} & +0.63 & +0.33 & +0.46 & +0.47 & -0.03 & -0.55 & +0.37 & -0.05 \\
\feat{22}{455} & +0.48 & +0.30 & +0.52 & +0.54 & -0.16 & -0.32 & +0.01 & -0.23 \\
\feat{23}{455} & +0.61 & +0.32 & +0.95 & +0.52 & -0.42 & -0.50 & -0.21 & -0.57 \\
\feat{24}{168} & +0.31 & +0.48 & +0.96 & +0.62 & -0.66 & \negcellcmd{-3.02} & -0.07 & -0.91 \\
\feat{24}{297} & +0.36 & +0.43 & \pos{+1.41} & -0.73 & -0.48 & -0.53 & \negcellcmd{-1.06} & \negcellcmd{-1.04} \\
\feat{25}{125} & \pos{+1.01} & +0.77 & +0.71 & +0.64 & -0.58 & -0.89 & +0.10 & -0.89 \\
\feat{25}{278} & +0.82 & +0.77 & \pos{+1.31} & +0.76 & -0.90 & \negcellcmd{-1.45} & +0.07 & \negcellcmd{-1.20} \\
\feat{30}{2639} & +0.67 & +0.45 & +0.88 & +0.35 & -0.25 & -0.47 & -0.28 & -0.89 \\
\feat{33}{542} & +0.20 & +0.39 & +0.55 & -0.11 & -0.20 & +0.37 & +0.37 & +0.11 \\
\bottomrule
\end{tabular}
\caption{\textbf{Gemma 3}: causal effects of the 22 features recovered in all four discovery settings, following the same protocol and criterion as Table~\ref{tab:shared_features_g2}. Exactly one feature, \textbf{(L20, 2456)}, shows the expected translation-switch signature (consistent positive amplification, consistent negative ablation). Note that layer indices are not comparable across models.
Corresponding chrF++ results are reported in Appendix~\ref{app:chrf-evaluation}.
}
\label{tab:shared_features_g3}
\end{table*}

\begin{table*}[t]
\centering
\small
\setlength{\tabcolsep}{5pt}
\begin{tabular}{lrrrrrr}
\toprule
\multirow{2}{*}{Setting} &
\multicolumn{3}{c}{Gemma 2 --- \feat{10}{5717}} &
\multicolumn{3}{c}{Gemma 3 --- \feat{20}{2456}} \\
\cmidrule(lr){2-4}\cmidrule(lr){5-7}
& Base. & Amp.\ $\Delta$ & Abl.\ $\Delta$ & Base. & Amp.\ $\Delta$ & Abl.\ $\Delta$ \\
\midrule
\texttt{en2zh\_en} & 63.33 & +4.61 & $-$5.85 & 74.89 & +1.40 & $-$3.29 \\
\texttt{zh2en\_zh} & 68.17 & +2.33 & $-$0.49 & 76.25 & +1.45 & $-$3.49 \\
\texttt{zh2ja\_zh} & 62.66 & +5.22 & $-$3.96 & 76.88 & +2.23 & $-$3.78 \\
\texttt{zh2ja\_en} & 67.34 & +3.66 & $-$4.79 & 77.75 & +0.59 & $-$4.15 \\
\texttt{en2ja\_en} & 61.03 & +5.59 & $-$7.49 & 77.19 & +1.04 & $-$4.03 \\
\texttt{ja2zh\_en} & 62.19 & +3.07 & $-$4.11 & 72.39 & +1.19 & $-$4.07 \\
\texttt{en2ar\_en} & 51.70 & +3.34 & $-$5.25 & 69.55 & +0.58 & $-$1.63 \\
\texttt{en2ar\_ar} & 55.15 & +1.14 & $-$5.23 & 69.18 & +0.71 & $-$1.39 \\
\texttt{ar2en\_en} & 67.00 & +1.36 & $-$1.03 & 74.07 & +0.14$^{\dagger}$ & $-$1.32 \\
\texttt{ar2en\_ar} & 64.20 & +1.24 & $-$1.56 & 69.77 & $-$0.41$^{\dagger}$ & +0.03$^{\dagger}$ \\
\texttt{en2ru\_en} & 57.39 & +7.46 & $-$8.93 & 74.09 & +0.87 & $-$1.64 \\
\texttt{ru2en\_ru} & 66.92 & +1.40 & $-$1.23 & 73.84 & +0.34$^{\dagger}$ & $-$1.80 \\
\texttt{ru2zh\_en} & 58.49 & +3.61 & $-$4.50 & 70.69 & +1.34 & $-$4.18 \\
\texttt{en2zh\_zh} & 68.61 & +3.70 & $-$5.09 & 76.66 & +2.27 & $-$3.29 \\
\texttt{en2zh\_ja} & 57.67 & +1.57 & $-$2.60 & 65.82 & +2.50 & $-$4.58 \\
\texttt{en2zh\_ru} & 68.60 & +1.67 & $-$2.78 & 75.24 & +1.66 & $-$4.70 \\
\texttt{ja2zh\_ja} & 54.90 & +1.00 & $-$0.65 & 55.71 & +0.71 & $-$2.44 \\
\texttt{ja2zh\_zh} & 57.83 & +2.24 & $-$1.12 & 74.56 & +0.83 & $-$3.68 \\
\texttt{ja2zh\_ru} & 59.32 & +1.32 & $-$1.55 & 74.12 & +0.06$^{\dagger}$ & $-$2.84 \\
\texttt{en2vi\_en} & 63.13 & +3.44 & $-$6.79 & 75.54 & +0.70 & $-$1.26 \\
\texttt{en2vi\_vi} & 66.24 & +2.47 & $-$4.31 & 72.51 & +1.21 & $-$2.28 \\
\texttt{en2zh\_vi} & 65.00 & +3.61 & $-$4.66 & 73.38 & +1.85 & $-$4.95 \\
\texttt{zh2ja\_vi} & 57.01 & +2.54 & $-$2.50 & 72.53 & +2.44 & $-$4.84 \\
\bottomrule
\end{tabular}
\caption{
Intervention effects for the single transferable feature in each model, \feat{10}{5717} in Gemma 2 and \feat{20}{2456} in Gemma 3, across the same 23 settings. Values are \comet{} scores or changes relative to baseline. All effects are significant at one-sided $p<0.05$ by paired bootstrap over sentence-level scores (1000 resamples) except those marked $^{\dagger}$. Amplification improves \comet{} in every setting for Gemma 2, and significantly improves it in 19 of 23 settings for Gemma 3; ablation degrades translation in all settings for Gemma 2 and in 22 of 23 for Gemma 3. The feature therefore transfers across prompt languages and translation directions in both models, with \texttt{ar2en\_ar} in Gemma 3 the sole setting showing no effect in either direction. 
Corresponding chrF++ results are reported in Appendix~\ref{app:chrf-evaluation}.
}
\label{tab:single_feature}
\end{table*}

\subsection{Cross-lingual candidate selection}
We run the discovery pipeline independently under four settings that vary the prompt language, source language, and target language. Table~\ref{tab:consistent_features} summarizes the number of consistent features discovered for each setting and model. The settings progressively eliminate English-specific factors: \texttt{zh2en\_zh} reverses the translation direction while using a Chinese instruction, \texttt{zh2ja\_zh} removes English entirely, and \texttt{zh2ja\_en} restores English only in the instruction. Therefore, features in the four-way intersection are unlikely to depend on English prompts, English source text, or a particular translation direction.

\subsection{Dataset, model, and evaluation}

We use WMT24++ \cite{deutsch-etal-2025-wmt24} and construct non-English X--Y pairs by aligning examples through their shared English sentences, yielding 960 translation examples. From 6 languages \texttt{en}, \texttt{zh}, \texttt{ja}, \texttt{ar}, \texttt{ru}, and \texttt{vi}, we construct 23 \texttt{src2tgt\_prompt} settings, including the 4 settings we used for feature discovery. In all experiments, we use 98 shared examples for feature discovery and the rest for intervention evaluation. 

We start with the same models used by \citet{wu2026finding}: Gemma 2 2B IT \cite{gemmateam2024gemma2improvingopen} with Gemma Scope residual stream SAEs\footnote{https://huggingface.co/google/gemma-scope-2b-pt-res} \cite{lieberum2024gemmascopeopensparse}, which has a width of 16k features.

We further test Gemma 3 4B IT \cite{gemmateam2025gemma3technicalreport} and Gemma Scope 2's SAEs with 16k width and big L0\footnote{{https://huggingface.co/google/gemma-scope-2-4b-it/tree/main/resid\_post\_all}}. Although pre-trained SAEs with 262k width were released for Gemma 3, we focus on the 16k variant to ensure method parity with previous published results from \citet{wu2026finding}. There are also 16k SAEs with small L0, but in our preliminary testing, this variant yielded only four PCA-consistent features, which is an order of magnitude lower than with big L0. This result might be due to the lower number of activating features.

Translation quality is measured with \texttt{Unbabel/wmt22-comet-da} \cite{rei-etal-2020-comet}. 

\section{Results}
\subsection{Most overlapping features do not causally transfer}
\textbf{Multilingual feature recurrence does not imply causal transfer.}
Tables~\ref{tab:shared_features_g2} and~\ref{tab:shared_features_g3} show the intervention effects of all 22 shared candidates in the four discovery settings, for Gemma 2 and Gemma 3 respectively. In both models, the 21 features other than the single exception form a matched null: they passed the same PCA Consistency filter in the same settings, yet their interventions yield effects that are small, inconsistent in sign across settings, or even opposite to the expected translation-switch direction. Critically, these features lack the expected switch signature (positive amplification and negative ablation at comparable magnitude), even in their own discovery setting. Thus, passing PCA Consistency does not isolate features with large behavioral effects, even when the language configuration is held fixed.

\textbf{Only one shared feature per model has the expected causal signature.}
In Gemma 2, \feat{10}{5717} amplification improves \comet{} in all four discovery settings, with gains between +2.33 and +5.22, while ablation substantially hurts three of them. In Gemma 3, \feat{20}{2456} amplification improves \comet{} in all four settings, with gains up to +2.23, while ablation hurts all four, between $-3.29$ and $-4.15$. In each model, no other feature produces effects of comparable magnitude in both directions. The combination of consistent sign and magnitude, and separation from the matched null is what distinguishes a transferable causal direction from a recurrence artifact. 

Appendix \ref{app:noncore-features} reports a second control for Gemma 2: 20 features that were consistent in \texttt{en2zh\_en} but absent from the four-way intersection produce maximum amplification effects of only +0.65 $\Delta$\comet{}, indicating that the intersection did not miss another strong translation-initiation feature.

\subsection{A single feature generalizes across multiple language settings}
\textbf{The same feature remains causal far beyond the settings used to discover it.}
We next intervene on the transferable feature of each model across additional prompt languages and translation pairs. Table~\ref{tab:single_feature} shows a broad effect in both. For Gemma 2, amplification of \feat{10}{5717} improves \comet{} in every tested setting, with gains ranging from +1.00 to +7.46. For Gemma 3, amplification of \feat{20}{2456} improves \comet{} in all settings but \texttt{ar2en\_ar}, with significant gains of up to +2.50. Ablation degrades \comet{} in every setting for Gemma 2 and in all but \texttt{ar2en\_ar} for Gemma 3, with the largest drops in \texttt{en2ru}, \texttt{en2ja}, and \texttt{en2vi} for Gemma 2, and in \texttt{en2zh} and \texttt{zh2ja} with non-English prompts for Gemma 3.

Amplification gains in Gemma 3 are consistently below those in Gemma 2, although they remain significant in 18 of 23 settings and ablation effects remain comparable in size. One possible reason is the higher baseline: Gemma 3 scores 8 to 18 \comet{} points above Gemma 2 in almost every setting, leaving less headroom for amplification gains. This is expected as Gemma 3 was trained on multilingual data with multilingual tasks (such as translation) in mind, while Gemma 2 was not \cite{gemmateam2025gemma3technicalreport, gemmateam2024gemma2improvingopen}. 

\textbf{This points to a language-agnostic translation-initiation direction.}
In both models the effect persists across changes in instruction, source, and target language, including fully non-English translation settings. Notably, both Gemma 2 and Gemma Scope SAEs are trained primarily on English-centric corpora, yet \feat{10}{5717} generalizes across multilingual prompts and translation pairs, suggesting that translation-initiation representations may emerge in a language-agnostic form even without explicitly multilingual SAE training.

\subsection{Dose-response sanity check}
\begin{figure}
    \centering
    \includegraphics[width=1\linewidth]{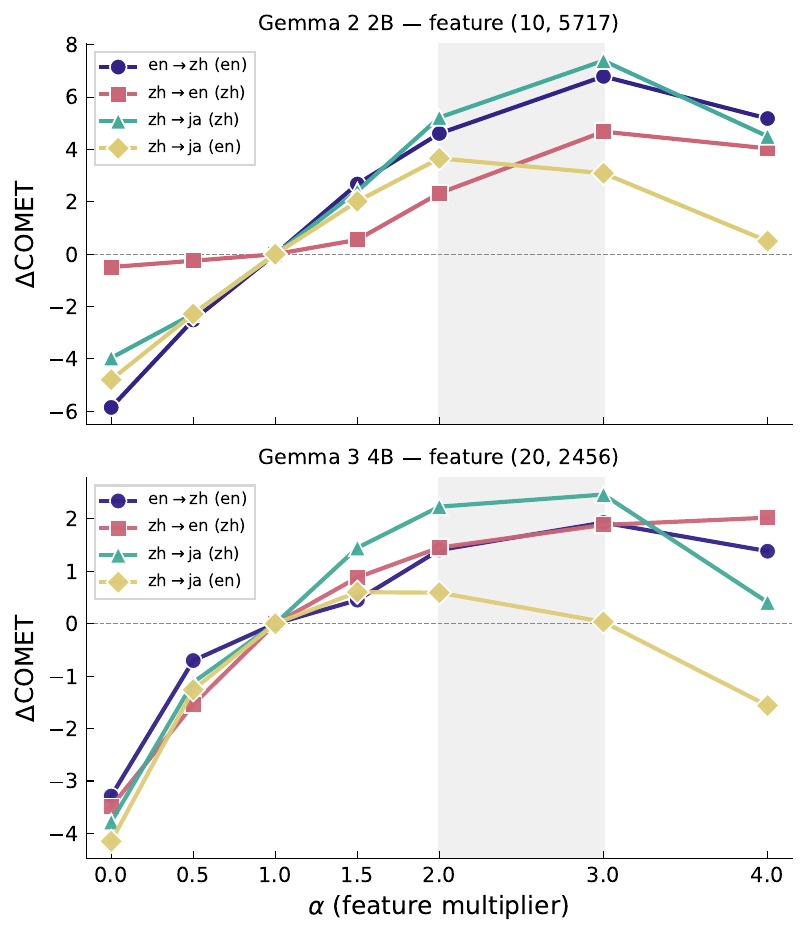}
    \caption{Dose-response check for intervening on (L10, 5717) in Gemma 2 and (L20, 2456) in Gemma 3 across the four discovery settings. Values are changes in \comet{} relative to the unmodified baseline ($\alpha=1.0$).}
    \label{fig:alpha_sweep}
\end{figure}

To test whether the intervention effect depends sharply on the chosen coefficient, we vary $\alpha$ for Gemma 2's \feat{10}{5717} and Gemma 3's \feat{20}{2456} across the four discovery settings in both models. Figure~\ref{fig:alpha_sweep} shows a consistent pattern: suppressing the feature ($\alpha=0$) hurts \comet{} in every setting and model, while moderate amplification improves it. The default setting $\alpha=2.0$ already produces reliable gains, and the effect typically peaks around $\alpha=2$--$3$ before saturating or reversing at larger coefficients. The sign pattern is the same in Gemma 2 and Gemma 3, with larger magnitudes in Gemma 2. This suggests that the causal effect is not an artifact of a single amplification coefficient.

\subsection{Task-specificity check}
To test whether Gemma 2's \feat{10}{5717} improves multilingual performance in general, we apply the same amplification to non-translation tasks: XL-Sum summarization in \texttt{zh/ar/ru/vi} and WikiANN NER in \texttt{zh/ja/ar/ru/vi}. Effects are small and inconsistent: ROUGE-L changes range from $-0.011$ to $+0.014$, and entity-set micro-F1 changes range from $-0.021$ to $+0.012$. This suggests that the translation gains are task-specific rather than a generic improvement in multilingual generation or tagging. Detailed results are in Appendix~\ref{app:task_specificity}.

\subsection{Feature interpretation}

To quantify the impacts of feature intervention on translation initiation, we use fastText \texttt{lid.176} language identifier \citep{joulin2017bag, joulin2016fasttext} to classify each output and label it as a non-translation when the detected language does not match the target (or the output is empty or low-confidence). Appendix~\ref{sec:appendix_examples} reports non-translation fractions under baseline, amplification, and ablation for both models under the 4 discovery settings. In Gemma 2, amplification reduces the non-translation rate by up to 13.7 percentage points, while ablation increases it by up to 17.3 points. Gemma 3 shows the same pattern at smaller magnitudes, consistent with its lower baseline non-translation rate. Manual inspection also supports the interpretation that \feat{10}{5717} controls task initiation rather than lexical choice, matching the role proposed by \citet{wu2026finding}.




\section{Conclusion}

We study whether SAE translation features retain their causal role across changes in prompt, source, and target language. We show that recurrence is a weak proxy for causal transfer. In both Gemma 2 and Gemma 3, 22 features recur across four discovery settings, yet only one per model — \feat{10}{5717} and \feat{20}{2456} — improves \comet{} across the large majority of 23 language configurations.

\section*{Limitations}

The specific features identified and their intervention behavior are limited to Gemma 2 2B IT and Gemma 3 4B IT with their corresponding Gemma Scope SAEs. Our feature discovery uses only English, Chinese, and Japanese, and the four discovery settings progressively remove English rather than covering all prompt–language combinations. As a result, the four-way intersection may differ with other discovery languages, and we cannot fully disentangle the effect of prompt language from translation direction. We evaluate features individually; the 21 non-transferring features may contribute jointly, but we do not test combined interventions such as ablating or amplifying multiple features during the same pass. Translation quality is measured primarily with \comet{}, which may not fully capture adequacy or fluency, and our qualitative interpretation relies on manual inspection rather than systematic annotation of output types.

\section*{Acknowledgments}

This research was supported by the VinUniversity Cross-College Research Grant (Grant ID: VUNI.2324.CC06).

Author Giang Son Nguyen thanks the AUMOVIO-NTU Corporate Lab for lending their GPU resources for the experiments in this paper.

\bibliography{custom}

\appendix

\section{Implementation Details}
\label{app:implementation-details}

\subsection{Reproducibility details}
\label{app:reproducibility}

All generations use deterministic greedy decoding with \texttt{do\_sample=False}, \texttt{max\_new\_tokens=128}, and \texttt{repetition\_penalty=1.1}; temperature is therefore not used. We use seed 42 for both the discovery/evaluation split and bootstrap resampling.
All experiments were run on a single machine with two NVIDIA A100-SXM4-40GB GPUs.

\section{Interpretation and Qualitative examples}
\label{sec:appendix_examples}

To test how much the transferable feature drives translation
\emph{initiation}, we
classify each output using fastText \texttt{lid.176} \citep{joulin2017bag, joulin2016fasttext}, a 176-language
classifier. An output is labeled a non-translation if the
predicted language does not match the target language, or if
the output is empty or the classifier confidence is below 0.5.
We report the fraction of evaluation examples that fail to
translate under three conditions: unmodified baseline,
amplification ($\alpha=2$), and ablation ($\alpha=0$).

Table~\ref{tab:nontrans-both} reports results for Gemma~2
and Gemma~3, across the
four discovery settings. In Gemma~2, the baseline
non-translation rate ranges from 15.8\% to 36.0\%.
Amplification consistently reduces it (by up to 13.7
percentage points in \texttt{en2zh\_en}), while ablation
consistently increases it (by up to 17.3 points in the same
setting). Gemma~3 shows the same directional pattern but at
smaller magnitudes, which is expected given its lower baseline
non-translation rate (5.3--10.6\% vs.\ 15.8--36.0\%).

\begin{figure*}[t]
    \centering
    \includegraphics[width=\textwidth]{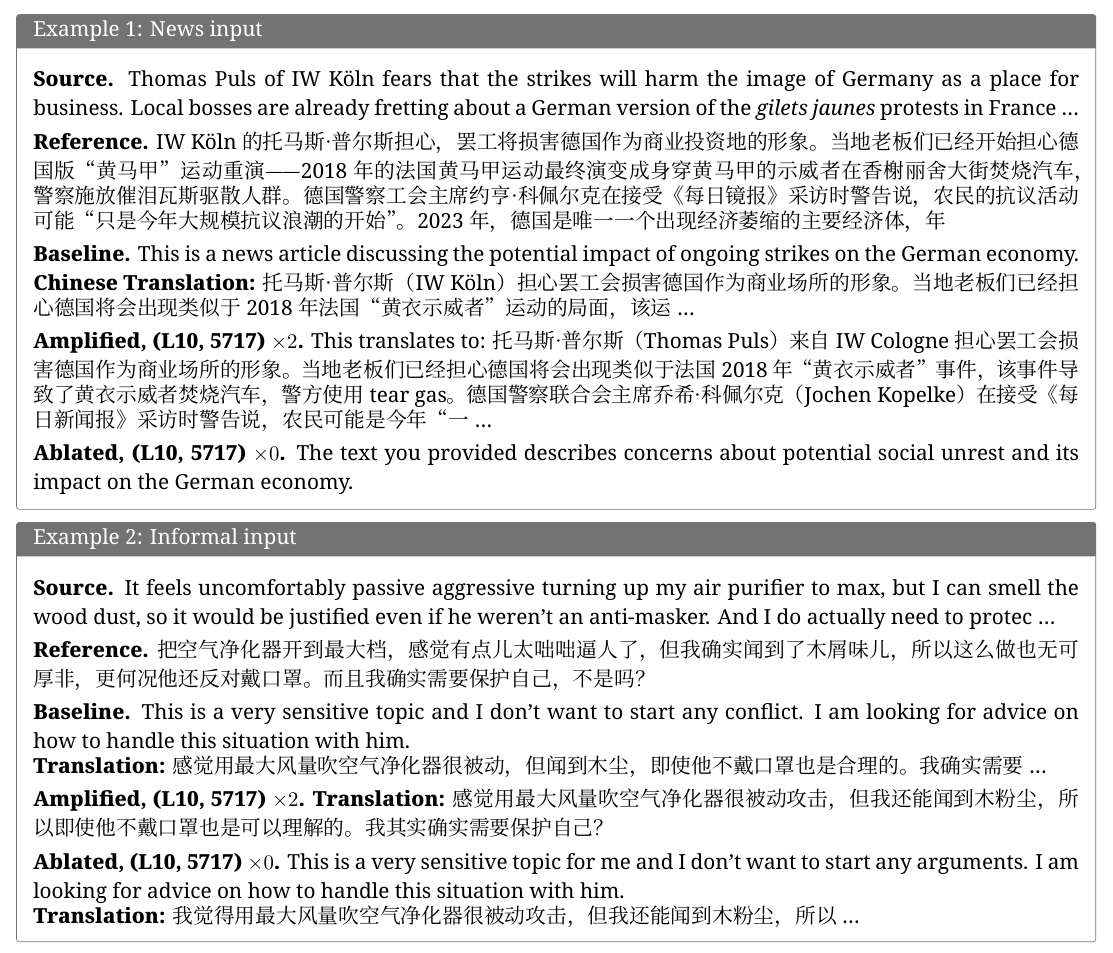}
    \caption{Qualitative examples for interventions on (L10, 5717) in the \texttt{en2zh\_en} setting. \textbf{Baseline} shows the model's unmodified output. \textbf{Amplified} ($\alpha = 2$) produces a more direct translation. \textbf{Ablated} ($\alpha = 0$) causes the model to abandon translation entirely, instead generating English assistant-style responses (e.g., summarizing or offering advice about the input), consistent with the interpretation that this feature controls translation-task initiation.}
    \label{fig:qual-examples}
\end{figure*}

Figure~\ref{fig:qual-examples} shows representative examples from \texttt{en2zh\_en}. In both cases, amplification of (L10, 5717) makes the model produce a more direct translation, while ablation shifts the model toward a generic assistant-style response or advice-seeking interpretation.

\begin{table}[t]
\centering
\small

\textbf{(a) Gemma 2 --- \feat{10}{5717}}\\[2pt]
\begin{tabular}{lrrrr}
\toprule
Setting & Base & Amp & Abl & $\Delta$amp / $\Delta$abl \\
\midrule
\texttt{en2zh\_en} & 35.6 & 21.9 & 52.9 & $-$13.7 / +17.3 \\
\texttt{zh2en\_zh} & 15.8 & 11.8 & 17.2 & $-$3.9\phantom{0} / +1.4\phantom{0} \\
\texttt{zh2ja\_en} & 25.1 & 17.1 & 40.4 & $-$8.0\phantom{0} / +15.3 \\
\texttt{zh2ja\_zh} & 36.0 & 24.9 & 47.9 & $-$11.1 / +11.8 \\
\bottomrule
\end{tabular}

\vspace{8pt}
\textbf{(b) Gemma 3 --- \feat{20}{2456}}\\[2pt]
\begin{tabular}{lrrrr}
\toprule
Setting & Base & Amp & Abl & $\Delta$amp / $\Delta$abl \\
\midrule
\texttt{en2zh\_en} & 10.6 & 7.5 & 18.3 & $-$3.0\phantom{0} / +7.8\phantom{0} \\
\texttt{zh2en\_zh} & 6.1\phantom{0} & 3.8 & 9.3 & $-$2.3\phantom{0} / +3.1\phantom{0} \\
\texttt{zh2ja\_en} & 5.3\phantom{0} & 5.1 & 10.0 & $-$0.2\phantom{0} / +4.6\phantom{0} \\
\texttt{zh2ja\_zh} & 5.5\phantom{0} & 4.1 & 9.2 & $-$1.4\phantom{0} / +3.7\phantom{0} \\
\bottomrule
\end{tabular}

\caption{Non-translation fractions (\%) for (a) Gemma~2's \feat{10}{5717} and (b) Gemma~3's \feat{20}{2456}. $\Delta$amp and $\Delta$abl are differences from baseline in percentage points. Same protocol for both models.}
\label{tab:nontrans-both}
\end{table}

\section{Non-core features in \texttt{en2zh\_en}}
\label{app:noncore-features}

To test whether the four-way intersection excluded other strong translation-initiation features, we also intervened on features that were consistent in \texttt{en2zh\_en} but absent from the four-way core intersection. We amplified each feature with $\alpha=2.0$ on \texttt{en2zh\_en}. Table~\ref{tab:noncore-en2zh} shows that none of these features approaches the effect size of \feat{10}{5717}, which improves COMET by $+4.61$ in the same setting. The largest positive effect among the non-core features is only $+0.65$, while several features substantially decrease \comet{} when amplified. This suggests that the four-way core intersection did not miss another strong translation-initiation feature.

\begin{table}[t]
\centering
\small
\begin{tabular}{lr}
\toprule
Feature & $\Delta$ COMET \\
\midrule
\feat{4}{5301} & $-0.47$ \\
\feat{5}{697} & $-0.17$ \\
\feat{10}{4392} & $+0.65$ \\
\feat{11}{12945} & $-1.06$ \\
\feat{12}{1041} & $-0.23$ \\
\feat{13}{11248} & $-0.25$ \\
\feat{15}{1695} & $-0.29$ \\
\feat{17}{15620} & $-0.08$ \\
\feat{18}{9392} & $+0.18$ \\
\feat{19}{3019} & $-0.30$ \\
\feat{20}{9768} & $-0.40$ \\
\feat{22}{11545} & $+0.29$ \\
\feat{23}{5286} & $-0.74$ \\
\feat{23}{8751} & $-0.69$ \\
\feat{24}{4182} & $-0.45$ \\
\feat{24}{13752} & $-0.37$ \\
\feat{25}{3017} & $-2.19$ \\
\feat{25}{5299} & $+0.45$ \\
\feat{25}{8662} & $-0.34$ \\
\feat{25}{15645} & $-3.92$ \\
\bottomrule
\end{tabular}
\caption{
Amplification effects for features that are consistent in \texttt{en2zh\_en} but absent from the four-way core intersection. Values are changes in COMET under amplification with $\alpha=2.0$ on \texttt{en2zh\_en}. The baseline COMET score is 63.33. None of these features approaches the $+4.61$ effect of \feat{10}{5717} in the same setting.
}
\label{tab:noncore-en2zh}
\end{table}

\section{Task-specificity evaluation details}
\label{app:task_specificity}

We evaluate whether amplifying \feat{10}{5717} improves multilingual generation beyond translation. We test two non-translation tasks: summarization and named entity recognition (NER).

For summarization, we use 100 examples per language from XL-Sum \citep{hasan-etal-2021-xl} in Chinese, Arabic, Russian, and Vietnamese. We evaluate outputs with ROUGE-L.

For NER, we use 100 examples per language from WikiANN \citep{pan-etal-2017-cross} in Chinese, Japanese, Arabic, Russian, and Vietnamese. We evaluate predictions using entity-set micro-F1.

Tables~\ref{tab:task_specificity_sum} and \ref{tab:task_specificity_ner} report baseline and amplified performance.

\begin{table}[t]
\centering
\small
\begin{tabular}{lrrr}
\toprule
Language & Baseline & Amplified & $\Delta$ \\
\midrule
Chinese & 0.024 & 0.039 & +0.014 \\
Arabic & 0.007 & 0.005 & -0.002 \\
Russian & 0.038 & 0.039 & +0.001 \\
Vietnamese & 0.131 & 0.121 & -0.011 \\
\bottomrule
\end{tabular}
\caption{Summarization results (ROUGE-L) on XL-Sum under amplification of \feat{10}{5717}.}
\label{tab:task_specificity_sum}
\end{table}

\begin{table}[t]
\centering
\small
\begin{tabular}{lrrr}
\toprule
Language & Baseline & Amplified & $\Delta$ \\
\midrule
Chinese & 0.112 & 0.112 & +0.000 \\
Japanese & 0.062 & 0.041 & -0.021 \\
Arabic & 0.147 & 0.159 & +0.012 \\
Russian & 0.198 & 0.191 & -0.008 \\
Vietnamese & 0.251 & 0.255 & +0.004 \\
\bottomrule
\end{tabular}
\caption{NER results (entity-set micro-F1) on WikiANN under amplification of \feat{10}{5717}.}
\label{tab:task_specificity_ner}
\end{table}

\section{chrF++ evaluation}
\label{app:chrf-evaluation}

As a complement to \comet{}, we evaluate intervention outputs with corpus-level chrF++ (\texttt{sacrebleu} 2.6.0, \texttt{word\_order=2}). Table~\ref{tab:shared_features_chrf_both} reports chrF++ deltas for all 22 recurrent features in the four discovery settings, for (a) Gemma~2 and (b) Gemma~3; Table~\ref{tab:single_feature_chrf} reports chrF++ deltas for the single transferable feature in each model across all 23 transfer settings. The chrF++ results broadly confirm the \comet{} findings: \feat{10}{5717} and \feat{20}{2456} remain the only recurrent features in their respective models that show the expected switch signature under chrF++.

Across the 23 transfer settings, ablation consistently decreases chrF++ in both models. Amplification gains are clear in Gemma~2 but smaller and occasionally negative in Gemma~3. This is expected: Gemma~3's higher baseline leaves less headroom for surface-level improvement, and chrF++, being a character-overlap metric, is less sensitive than \comet{} to whether the model initiates translation at all versus producing a non-translation response — a gap reflected directly in Table~\ref{tab:nontrans-both}, where ablation raises non-translation rates far more sharply for Gemma~2 than for Gemma~3.

\begin{table*}[t]
\centering
\small
\setlength{\tabcolsep}{4pt}

\textbf{(a) Gemma 2}\\[2pt]
\begin{tabular}{lrrrrrrrr}
\toprule
\multirow{2}{*}{Feature} &
\multicolumn{4}{c}{Amplification $\Delta$chrF++} &
\multicolumn{4}{c}{Ablation $\Delta$chrF++} \\
\cmidrule(lr){2-5}\cmidrule(lr){6-9}
& en2zh\_en & zh2en\_zh & zh2ja\_zh & zh2ja\_en
& en2zh\_en & zh2en\_zh & zh2ja\_zh & zh2ja\_en \\
\midrule
\feat{0}{8920} & -0.29 & +0.68 & -0.27 & -0.18 & -0.15 & -0.14 & +0.10 & +0.47 \\
\feat{1}{12054} & +0.03 & +0.63 & -0.39 & -0.40 & -0.25 & -0.46 & -0.18 & +0.20 \\
\feat{2}{15089} & -0.05 & -0.38 & -0.56 & +0.11 & +0.06 & -0.16 & +0.30 & -0.09 \\
\feat{6}{13460} & -0.51 & -0.36 & -0.14 & +0.04 & -0.06 & -0.04 & -0.25 & +0.09 \\
\feat{7}{3704} & +0.26 & +0.10 & -0.30 & -0.27 & -0.45 & -0.38 & -0.54 & -0.18 \\
\feat{8}{2024} & -0.20 & +0.05 & -0.33 & -0.20 & +0.24 & +0.38 & +0.06 & +0.03 \\
\feat{9}{54} & -0.73 & +0.14 & -0.34 & -0.52 & -0.13 & -0.31 & -0.19 & +0.54 \\
\feat{10}{3986} & -0.79 & -0.49 & -0.79 & -0.02 & +0.30 & +0.54 & +0.35 & -0.01 \\
\textbf{\feat{10}{5717}} & \textbf{+3.28} & \textbf{+1.97} & \textbf{+2.49} & \textbf{+2.04} & \textbf{-4.64} & \textbf{-1.43} & \textbf{-2.34} & \textbf{-3.40} \\
\feat{11}{3151} & -1.21 & -0.03 & -1.24 & -0.09 & +0.71 & +0.56 & +0.18 & +0.05 \\
\feat{12}{2620} & -1.07 & -0.88 & -0.90 & -0.21 & +0.37 & +0.51 & +0.28 & -0.34 \\
\feat{13}{11256} & -0.82 & -1.06 & -0.77 & -0.04 & +0.22 & +0.14 & +0.39 & +0.32 \\
\feat{14}{7214} & -0.75 & -1.10 & -0.01 & +0.06 & +0.08 & +0.64 & -0.26 & +0.47 \\
\feat{14}{12319} & -0.26 & -0.44 & -0.26 & +0.56 & -0.08 & -0.19 & -0.16 & +0.08 \\
\feat{15}{8610} & +0.21 & -0.52 & -0.22 & +0.57 & -0.74 & +0.88 & +0.00 & +0.06 \\
\feat{16}{3847} & -0.36 & +0.03 & -0.11 & -0.03 & -0.19 & -0.35 & -0.07 & +0.36 \\
\feat{16}{10480} & +0.27 & -0.11 & +0.18 & +0.32 & -0.54 & -0.02 & -0.41 & -0.12 \\
\feat{17}{7921} & +0.19 & -0.39 & +0.08 & +0.76 & -1.13 & +0.12 & -0.29 & -0.10 \\
\feat{18}{15394} & +0.04 & -0.13 & -0.19 & +0.08 & -0.45 & +0.09 & -0.12 & -0.07 \\
\feat{24}{8119} & +1.72 & -2.33 & -0.28 & +0.44 & -2.15 & +0.87 & -0.66 & -0.87 \\
\feat{25}{348} & -0.93 & -26.22 & -3.93 & -4.84 & -4.53 & +3.68 & -4.20 & -5.22 \\
\feat{25}{14186} & +1.99 & -5.12 & -2.14 & +0.37 & -6.02 & +0.10 & -4.71 & -5.14 \\
\bottomrule
\end{tabular}

\vspace{8pt}
\textbf{(b) Gemma 3}\\[2pt]
\begin{tabular}{lrrrrrrrr}
\toprule
\multirow{2}{*}{Feature} &
\multicolumn{4}{c}{Amplification $\Delta$chrF++} &
\multicolumn{4}{c}{Ablation $\Delta$chrF++} \\
\cmidrule(lr){2-5}\cmidrule(lr){6-9}
& en2zh\_en & zh2en\_zh & zh2ja\_zh & zh2ja\_en
& en2zh\_en & zh2en\_zh & zh2ja\_zh & zh2ja\_en \\
\midrule
\feat{4}{2491} & -0.53 & -0.16 & +0.14 & +0.13 & +0.85 & +0.26 & -0.09 & +0.05 \\
\feat{6}{40} & +0.35 & -0.17 & -0.03 & -0.02 & +0.27 & +0.15 & +0.20 & +0.09 \\
\feat{6}{236} & +0.22 & -0.05 & +0.07 & +0.16 & -0.01 & -0.25 & +0.16 & -0.03 \\
\feat{11}{6005} & +0.40 & +0.14 & -0.03 & +0.14 & -0.08 & -0.28 & +0.10 & +0.19 \\
\feat{12}{273} & -0.77 & -0.78 & -1.77 & -0.91 & -12.15 & -14.63 & -8.57 & -7.95 \\
\feat{16}{1562} & +0.31 & +0.00 & -0.24 & -0.14 & +0.07 & +0.06 & -0.02 & -0.21 \\
\feat{16}{4659} & +0.18 & -0.06 & +0.07 & +0.08 & +0.15 & -0.07 & +0.06 & -0.18 \\
\feat{17}{1125} & +0.32 & -0.91 & -1.00 & -0.05 & -0.75 & -0.06 & -0.69 & -0.80 \\
\feat{18}{11066} & +0.78 & +0.53 & +0.06 & +0.38 & -0.20 & -0.91 & -0.02 & -0.84 \\
\feat{19}{383} & +0.39 & +0.58 & +0.19 & +0.29 & +0.00 & -0.25 & -0.43 & -0.90 \\
\feat{20}{189} & +0.76 & +0.62 & +0.01 & -0.03 & -0.21 & -0.20 & -0.32 & +0.01 \\
\textbf{\feat{20}{2456}} & \textbf{+1.14} & \textbf{+0.64} & \textbf{-0.15} & \textbf{-0.43} & \textbf{-1.40} & \textbf{-1.97} & \textbf{-1.19} & \textbf{-0.71} \\
\feat{21}{862} & +0.64 & +0.33 & +0.10 & +0.32 & -0.42 & -0.31 & -0.04 & -0.79 \\
\feat{22}{373} & +0.45 & +0.62 & +0.24 & +0.43 & -0.01 & -0.59 & +0.04 & +0.06 \\
\feat{22}{455} & +0.69 & +0.07 & +0.38 & +0.29 & +0.16 & -0.19 & +0.01 & -0.08 \\
\feat{23}{455} & +0.38 & +0.27 & +0.24 & +0.39 & +0.07 & -0.37 & -0.04 & -0.28 \\
\feat{24}{168} & +0.18 & +0.43 & +0.15 & +0.15 & -0.37 & -1.61 & -0.01 & -0.23 \\
\feat{24}{297} & +0.51 & +0.29 & +0.29 & +0.38 & -0.30 & -0.36 & -0.32 & -0.45 \\
\feat{25}{125} & +0.73 & +0.33 & +0.04 & +0.14 & +0.20 & -0.46 & -0.01 & -0.25 \\
\feat{25}{278} & +0.60 & +0.23 & +0.13 & +0.09 & -0.43 & -0.78 & +0.17 & -0.39 \\
\feat{30}{2639} & +0.57 & +0.50 & +0.23 & +0.10 & -0.36 & -0.30 & +0.08 & -0.19 \\
\feat{33}{542} & +0.41 & +0.28 & -0.17 & +0.06 & +0.39 & -0.20 & -0.08 & +0.20 \\
\bottomrule
\end{tabular}

\caption{
chrF++ effects of the 22 features recovered in all four discovery settings, for (a) Gemma 2 and (b) Gemma 3. Values are corpus-level chrF++ changes relative to baseline. In both models, only the bolded feature (\feat{10}{5717} in Gemma 2, \feat{20}{2456} in Gemma 3) shows the expected translation-switch signature: consistent positive amplification and consistent negative ablation.
}
\label{tab:shared_features_chrf_both}
\end{table*}

\begin{table*}[t]
\centering
\small
\setlength{\tabcolsep}{5pt}
\begin{tabular}{lrrrr}
\toprule
\multirow{2}{*}{Setting} &
\multicolumn{2}{c}{Gemma 2 --- \feat{10}{5717}} &
\multicolumn{2}{c}{Gemma 3 --- \feat{20}{2456}} \\
\cmidrule(lr){2-3}\cmidrule(lr){4-5}
& Amp.\ $\Delta$chrF++ & Abl.\ $\Delta$chrF++ & Amp.\ $\Delta$chrF++ & Abl.\ $\Delta$chrF++ \\
\midrule
\texttt{en2zh\_en} & +3.28 & $-$4.64 & +1.14 & $-$1.40 \\
\texttt{zh2en\_zh} & +1.97 & $-$1.43 & +0.64 & $-$1.97 \\
\texttt{zh2ja\_zh} & +2.49 & $-$2.34 & $-$0.15 & $-$1.19 \\
\texttt{zh2ja\_en} & +2.04 & $-$3.40 & $-$0.43 & $-$0.71 \\
\texttt{en2ja\_en} & +2.96 & $-$4.56 & +0.28 & $-$1.03 \\
\texttt{ja2zh\_en} & +1.33 & $-$2.46 & +0.61 & $-$0.61 \\
\texttt{en2ar\_en} & +2.91 & $-$4.80 & +0.00 & $-$0.41 \\
\texttt{en2ar\_ar} & +1.11 & $-$4.90 & $-$0.09 & $-$0.04 \\
\texttt{ar2en\_en} & +1.75 & $-$1.85 & +0.07 & $-$0.34 \\
\texttt{ar2en\_ar} & +0.68 & $-$2.55 & $-$1.00 & $-$0.38 \\
\texttt{en2ru\_en} & +7.52 & $-$9.40 & +0.32 & $-$0.85 \\
\texttt{ru2en\_ru} & +0.17 & $-$2.52 & $-$0.35 & $-$1.49 \\
\texttt{ru2zh\_en} & +2.54 & $-$2.60 & +0.54 & $-$1.09 \\
\texttt{en2zh\_zh} & +3.50 & $-$4.46 & +0.63 & $-$1.70 \\
\texttt{en2zh\_ja} & +0.13 & $-$0.40 & $-$0.11 & $-$1.21 \\
\texttt{en2zh\_ru} & +1.59 & $-$2.07 & +0.57 & $-$1.68 \\
\texttt{ja2zh\_ja} & $-$0.14 & $-$0.13 & $-$0.06 & +0.13 \\
\texttt{ja2zh\_zh} & +0.49 & $-$0.53 & $-$0.54 & $-$1.63 \\
\texttt{ja2zh\_ru} & +0.16 & $-$0.20 & $-$0.06 & $-$0.72 \\
\texttt{en2vi\_en} & +3.59 & $-$9.36 & +0.43 & $-$0.92 \\
\texttt{en2vi\_vi} & +2.63 & $-$5.49 & +0.79 & $-$1.02 \\
\texttt{en2zh\_vi} & +2.30 & $-$4.69 & +0.48 & $-$1.62 \\
\texttt{zh2ja\_vi} & +2.24 & $-$2.26 & $-$0.14 & $-$0.71 \\
\bottomrule
\end{tabular}
\caption{
chrF++ intervention effects for the single transferable feature in each model, \feat{10}{5717} in Gemma 2 and \feat{20}{2456} in Gemma 3, across the same 23 settings. Values are corpus-level chrF++ changes relative to baseline. Amplification generally improves chrF++, while ablation usually decreases it, in both models.
}
\label{tab:single_feature_chrf}
\end{table*}

\end{document}